\documentclass[conference]{IEEEtran}

\pdfoutput=1
\usepackage{cite}

\usepackage{hyperref}

\usepackage{graphicx}
\usepackage[hang]{subfigure}
\graphicspath{figures}
\usepackage{amsmath}
\usepackage{array}

\usepackage{caption}

\usepackage{url}

\usepackage{boxhandler}

\newcommand{\be}{\begin{equation}}
\newcommand{\ee}{\end{equation}}
\newcommand{\eat}[1]{}

\begin{document}
%
\title{Attention-based Extraction of Structured Information from Street View Imagery}

\author{
\IEEEauthorblockN{Zbigniew Wojna}
\IEEEauthorblockA{University College London \\
zbigniewwojna@gmail.com}
\and
\IEEEauthorblockN{Alex Gorban}
\IEEEauthorblockA{Google Inc.\\
gorban@google.com}
\and
\IEEEauthorblockN{Dary-Shyang Lee}
\IEEEauthorblockA{Google Inc. \\
dsl@google.com}
\and
\IEEEauthorblockN{Kevin Murphy}
\IEEEauthorblockA{Google Inc. \\
kpmurphy@google.com}
\and
\IEEEauthorblockN{Qian Yu}
\IEEEauthorblockA{Google Inc. \\
qyu@google.com}
\and
\IEEEauthorblockN{Yeqing Li}
\IEEEauthorblockA{Google Inc. \\
yeqing@google.com}
\and
\IEEEauthorblockN{Julian Ibarz}
\IEEEauthorblockA{Google Inc. \\
julianibarz@google.com}
}

%
\author{
\IEEEauthorblockN{
Zbigniew Wojna\IEEEauthorrefmark{1} \quad
Alex Gorban\IEEEauthorrefmark{2} \quad
Dar-Shyang Lee\IEEEauthorrefmark{2} \quad
Kevin Murphy\IEEEauthorrefmark{2} \\ 
Qian Yu\IEEEauthorrefmark{2} \quad
Yeqing Li\IEEEauthorrefmark{2} \quad
Julian Ibarz\IEEEauthorrefmark{2}
}
\\
\IEEEauthorblockA{\IEEEauthorrefmark{1} University College London}
\IEEEauthorblockA{\IEEEauthorrefmark{2} Google Inc.}
}


\maketitle

\begin{abstract}
  We present a neural network model --- based on Convolutional Neural Networks, Recurrent Neural Networks and
  a novel attention mechanism --- which achieves 84.2\% accuracy on the challenging French
Street Name Signs (FSNS) dataset, significantly outperforming the
previous state of the art (Smith'16), which achieved
72.46\%. Furthermore, our new method is much simpler and more general
than the previous approach. To demonstrate the generality of our
model, we show that it also performs well on
an  even more challenging dataset derived from
Google Street View,
in which the goal is to extract business names from
store fronts.
Finally, we study the speed/accuracy tradeoff that
results from using CNN feature extractors of different depths.
Surprisingly, we find that deeper is not always better (in terms of accuracy,
as well as speed).
Our resulting model is simple, accurate and fast, allowing it to be
used at scale on a variety of challenging real-world text extraction
problems.   
\end{abstract}

\section{Introduction}
\label{introduction}

Text recognition in an unconstrained natural environment is a
challenging computer vision and machine learning problem. Traditional Optical Character Recognition (OCR) systems mainly
focus on extracting text from scanned documents. Text acquired from
natural scenes is more challenging due to visual artifacts, such as
distortion, occlusions, directional blur, cluttered background or
different viewpoints. Despite these difficulties, recent advances in
deep learning have made significant progress on this problem
\cite{he2016text,wang2011end,Jaderberg2016,karatzas2015icdar,veit16cocotext,Iwamura2016}.     

In this paper, we concentrate not just on transcribing all the text in a
given image, but instead on the harder problem of extracting a subset
of useful pieces of
information. The model has to focus on the important parts of the
scene and to ignore visual clutter. We propose a model which
leverages convolutional neural networks (CNNs), recurrent neural
networks (RNNs), and a new form of spatial attention.

We benchmark our model on the French Street Name Signs dataset (FSNS)
\cite{smith2016end}, derived from Google Street View. The dataset
contains over 1M labeled images of 
visual text ``in the wild''; this is significantly more than COCO Text
\cite{veit16cocotext}, which only includes 63k labeled images. 
We achieve 84.2\% accuracy on FSNS,  significantly outperforming
the previous state-of-the-art \cite{smith2016end}, which achieved
72.46\%.

The previous state of the art method on FSNS
\cite{smith2016end} shifts different views of the same sign into the batch dimension, has multiple multilayer LSTMs designed to treat every line of text separately (up to 3 lines) and uses CTC loss. Our model is simpler, more accurate and makes fewer assumptions about the data.
To demonstrate its broad applicability, 
we also evaluate our new model on
the Street View Business
Names dataset \cite{yu2015large}, showing strong results.

Finally, we study the accuracy and speed of
using 3 different CNN-based feature extractors
(namely   inception-v2 \cite{szegedy2015rethinking},
inception-v3 \cite{szegedy2016inception} and inception-resnet-v2
\cite{szegedy2016inception})
as input to our attention model.
We find that inception-v3 and inception-resnet-v2
perform comparably, and both significantly outperform
inception-v2.
Motivated by the need for speed,
we also study the effect of using ``ablated'' versions of these models,
which use fewer layers.
Interestingly, we find that for all three networks,
the accuracy initially increases with depth,
but then starts to decrease.
  This is in contrast to models trained on the ILSVRC Imagenet dataset
  \cite{ILSVRCarxiv14}, which is comparable in size to FSNS.
For image classification, accuracy tends to increase with depth monotonically.
We believe the difference is that image classification needs very complicated features, which are spatially invariant, whereas, for text extraction, it hurts to use to use such features.

In summary, our contributions are as follows:
(1) We present a novel attention-based text reading architecture, trained
in an end-to-end way, that beats the previous state of the art on the FSNS dataset
by a significant margin while being considerably simpler and more
general. (2) We show how our new model also gives excellent results on
a newer, even more challenging,  Street View dataset. 
(3) We study the speed/accuracy tradeoff that results from
using CNNs of different depths and recommend a
configuration that is accurate and efficient. 
 
The source code and a trained model are available at:\\
https://github.com/tensorflow/models/tree/master/attention\_ocr

\section{Methods}

\begin{figure}
\begin{center}
\includegraphics[scale=0.8]{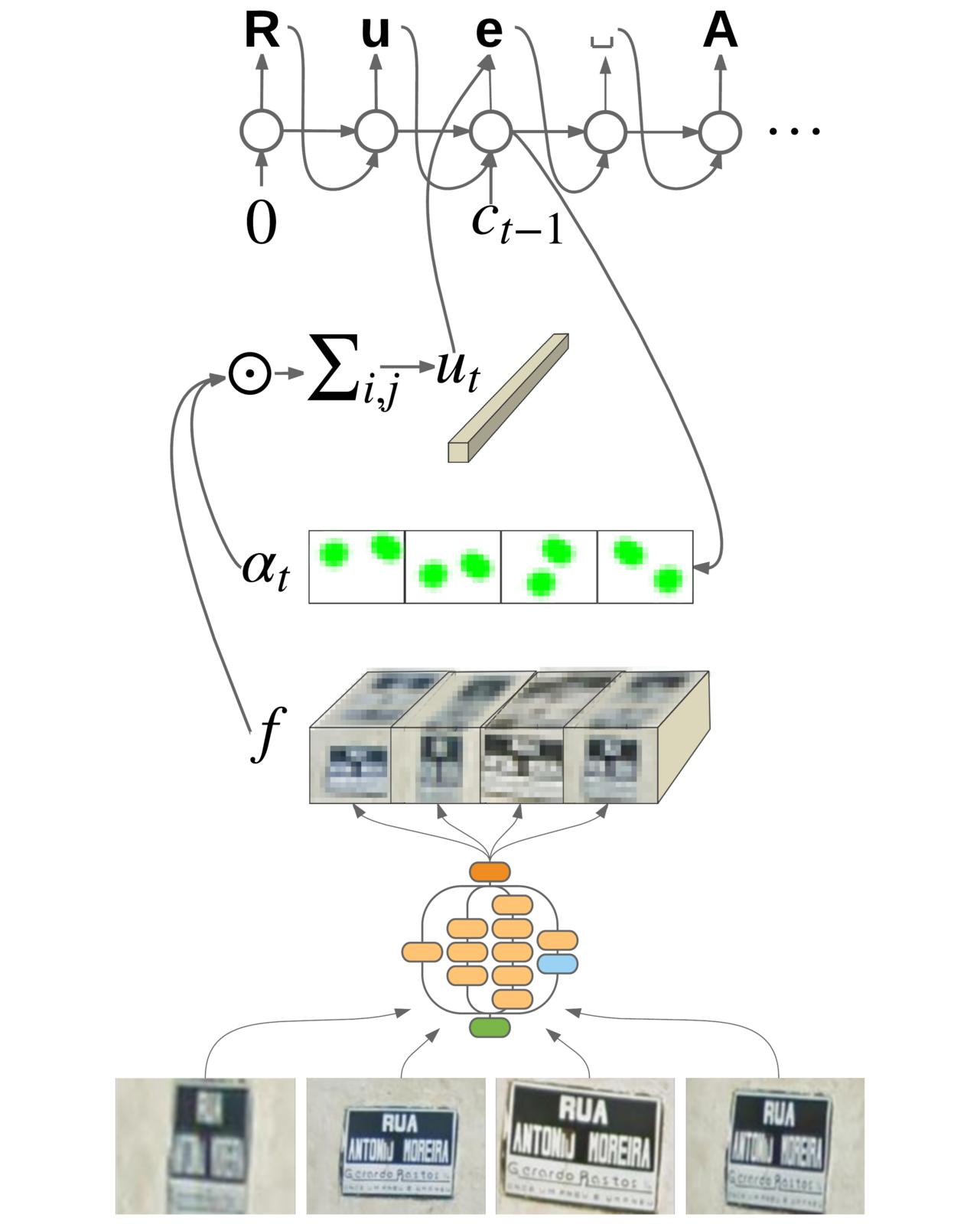}
\caption{Architecture of our model.
  We pass each of the four views through the same CNN feature extractor, and then concatenate the results into a single large feature map, shown by the cube labeled ``f''. We take a spatially weighted combination to create a fixed-sized feature vector $u_t$, which is fed into the RNN.
}
\label{fig:arch}
\end{center}
\end{figure}

In this section, we describe our model, which processes the image
through a CNN, and then passes the (attentionally weighted) features
into an RNN. See Figure~\ref{fig:arch} for a high level summary.  

\subsection{CNN-based feature extraction}

We consider 3 kinds of CNN: inception-v2 \cite{szegedy2015rethinking},
inception-v3 \cite{szegedy2016inception} and inception-resnet-v2
\cite{szegedy2016inception}, which combines inception
with resnets \cite{he2015deep}. These models achieve
state-of-the-art performance on the Imagenet classification challenge
\cite{ILSVRCarxiv14, canziani2016analysis}. 

It has been shown in \cite{yosinski2014transferable} that features
from lower layers of CNNs trained on Imagenet transfer well to other
tasks.
However, this still leaves open the question of which layer to use as
our feature representation. We study this in
Section~\ref{sec:cuts}.
(We also compared the effects of pre-training on Imagenet
vs. training from random initialization, and found no notable difference,
so we only report results using the latter.)

We will use $f= \{f_{i,j,c}\}$ to denote the feature map derived by
passing the image $x$ through a CNN (here $i,j$ index locations in the
feature map, and $c$ indexes channels).

\subsection{RNN}

The main challenge is how to convert the feature maps into a single
text string. Following previous work, we use an RNN
(specifically, an LSTM\cite{hochreiter1997long})
for this; this
acts as a character-level language model, which takes inputs from 
the image, as we explain below.

Let $s_t$ be the hidden state of the RNN at time $t$.
The input to the RNN is determined by a spatially weighted combination of image features.
This spatial attention mask is denoted by
$\alpha_t=\{\alpha_{t,i,j}\}$; we explain how to compute this in Section~\ref{sec:attn}.
Once we have computed the spatial mask, we
compute a weighted combination of the features (the context) as
follows:  
\be
u_{t,c} = \sum_{i,j} \alpha_{t,i,j} f_{i,j,c}
\ee
The total input to the RNN at time $t$ is defined as
\be
\hat{x}_{t} = W_c c_{t-1}^{OneHot} + W_{u_1} u_{t-1}
\ee
where $c_{t-1}$ is the index of the previous letter
(ground truth during training, predicted during test time).
We then compute the output and next state of the RNN as follows:
\be  
(o_t, s_t) = \mathrm{RNNstep}(\hat{x}_{t}, s_{t-1})
\ee
The final predicted distribution over letters at time $t$ is given by
\be
\hat{o}_t = \mathrm{softmax}(W_o o_t + W_{u_2} u_t)
\ee
This combines information from the RNN, $o_t$, with information from
the attentional feature vector, $u_t$.
Finally, we compute the most likely letter:
\be
c_t = \arg \max_c \hat{o}_t(c)
\ee
This is called greedy decoding.

\subsection{Spatial attention}
\label{sec:attn}

Most prior works that use spatial attention for OCR
(e.g., \cite{lee2016recursive,shi2015end,shi2016robust,he2016text, bluche2016scan, bluche2016joint})
predict the mask based on the current RNN state, as follows:
\be
a_{t,i,j} = V_a^T \tanh(W_s s_t + W_f f_{i,j,:})
\label{eqn:attn}
\ee
\be
\alpha_t = \mathrm{softmax}_{i,j}(a_t)
\label{eqn:attn2}
\ee
where $V_a$ is a vector and $\tanh$ is applied elementwise to its vector argument.
This combines content from the image, via $W_f f$, with a time-varying
offset, via $W_s s_t$, to determine where to look.
We will use this as our baseline attention method.

\begin{figure}
\begin{center}
\includegraphics[width=0.7\hsize]{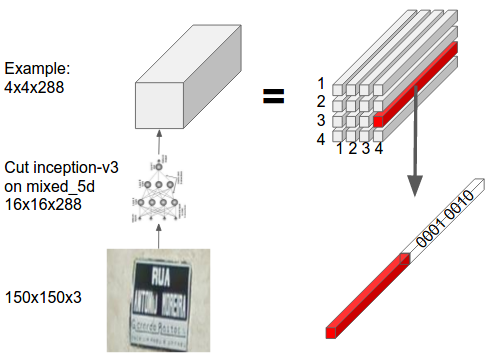}
\caption{Adding pixel coordinates to image features.}
\label{fig:coord}
\end{center}
\end{figure}

The above attention mechanism is permutation invariant, meaning we
could shuffle the order of the pixels and the mapping from $f$ to
$\alpha_t$ would remain the same (since it is applied elementwise to
each location). To make the model ``location aware'', we concatenate
$f_{i,j,:}$ with a one-hot encoding of the spatial coordinates
$(i,j)$,
as shown in  Figure~\ref{fig:coord}.
More precisely,
we replace the argument to the $\tanh$ function with the following:
\be
W_s s_t + W_{f_1} f_{i,j,:}  + W_{f_2}  e_i + W_{f_3}  e_j
\ee
where $e_i$ is a one-hot encoding of coordinate $i$,
and similarly for $e_j$.
This is equivalent to adding a spatially varying matrix of bias terms.

Our proposal is different than the location-aware attention mechanism
proposed in \cite{chorowski2015attention}. They suggested adding
$W_{a_2}*\alpha_{t-1}$ as input to the $\tanh$ function in
Equation~\ref{eqn:attn}, where $W_{a_2}$ is a convolution
kernel. However, in multiline text recognition problems, we sometimes
have to make a big jump to the left side of the line below
(see Figure~\ref{fig:bizname_heatmap} for example),
which cannot be captured by this approach.
(We tried their approach of location-aware attention, and it did not give good results for our problem, details omitted for brevity.)

\subsection{Handling multiple views}

In the FSNS dataset, we have four views for each input sign,
each of size 150x150. We process each of these independently, through
the same CNN-based feature extractor (parameters are shared), to
compute four feature maps.
We then concatenate these horizontally
to create a single input feature map.
For example, suppose the feature map for each
of the four views
is $16 \times 16 \times 320$;
then after concatenation, the feature map
$f_{i,j,c}$ will be $64 \times 16 \times 320$.
(The actual spatial resolution of the feature maps varies,
as we discuss in Section~\ref{sec:cuts}.)

\subsection{Training}

We train the model using (penalized) maximum likelihood estimation,
that is, we maximize $\sum_{t=1}^T \log p(y_t|y_{1:t-1},x)$, where $x$
is the input image, $y_t$ is the predicted label for location $t$, and
$T=37$ for the FSNS dataset. (If the output string is less than 37
characters, the model is required to predict a null character.)
Since the model is autoregressive,
we pass in the ground truth labels as history during training,
as is standard \cite{Sutskever2014}.
Note that we do not need ground truth bounding boxes around any of the
text, which makes collecting training data much easier.

Our training method
allows us to use autoregressive connections easily,
which is not possible  when using CTC loss \cite{graves2006connectionist},
which was used by 
the previous state of the art model on FSNS \cite{smith2016end}.
We find the use of such autoregressive dependencies
improves  accuracy by 6\%, and speeds up training 2x
(details omitted for brevity).

We use stochastic gradient descent optimization with initial learning
rate $0.002$, decay factor $0.1$ after $1,200,000$ steps and momentum
$0.75$. We finish training after $2,000,000$ steps. We use the following augmentation procedure per view: We
randomly crop with the requirement to cover at least $0.8$ area of the
original input image and new aspect ratio to be between $0.8$ and
$1.2$. After cropping we resize it to the initial size with one
randomly chosen interpolation procedure: bilinear, bicubic, area, or
nearest-neighbor. Then we apply random image distortions: change of
contrast, hue, brightness, and saturation. To regularize the model, we
use weight decay $0.00004$, label smoothing $0.9$ \cite{szegedy2015rethinking} and LSTM values
clipping to $10$. LSTM unit size is $256$. We use a batch size of $32$
with asynchronous training on $40$ machines. For inception-resnet-v2 we have
used a batch size of $12$ due to GPU memory limitations. It takes less than 10
hours to train a single model with the inception-v3 network. We apply Polyak averaging
\cite{polyak1992acceleration} with decay $0.9999$ to derive the
weights for inference. 

\section{Datasets}

In this section, we describe the datasets that we use in our experiments.

\subsection{FSNS dataset}

The FSNS dataset \cite{smith2016end} contains $965,917$ training
images, $38,633$ validation images and $42,839$ test images. Each
image has up to $4$ tiles, intended to be a different view of the same
physical street sign from Street View imagery from France. The size of
every tile is 150x150 pixels. The first view represents the ground
truth physical sign (as views do not always correspond to the same
sign). See Figure~\ref{fig:FSNSresults} for some examples.  

All the transcriptions of the street name are up to 37 characters
long. (Our model takes advantage of this fact and always runs 37
steps, with an optional out-of-alphabet padded symbol.) There are 134
possible characters to choose from at each location, but most of the
street names consist only of Latin letters. 

\subsection{Street View Business Names Dataset}

This is an internal dataset which contains $\sim 1M$ single view
images of business storefronts extracted from Google Street View 
imagery. See Figure~\ref{fig:bizname} for some examples. The size
of every image is 352x352. All transcriptions contain up to 33
symbols, with 128 characters in the vocabulary. 

This dataset is significantly more challenging than FSNS, as
storefronts have a lot more variation, and richer contextual
information, compared to the street name signs. Moreover, the business
name can be a small fraction of the entire image. 

\section{Experimental results}

In this section, we report our experimental results on various
datasets.

\subsection{Accuracy on FSNS}
\label{sec:accuracy}

We use the full sequence accuracy metric to benchmark our model. It is
a challenging metric, as it requires every character to agree with the
ground truth.  
We compare
the previous state-of-the-art \cite{smith2016end}
with five different versions of our model.
In particular, we use three feature extractors
(inception-v2, inception-v3 and inception-resnet-v2),
and combine this 
 with two attention
models (standard and  our novel position-dependent attention).
As we see from Table~\ref{tab:results},
all our methods significantly outperform the
previous state-of-the-art;
inception-v3 and inception-resnet-v2  give similar performance,
and both significantly beat
inception-v2;
finally, 
we see that our novel location-aware attention helps by 0.9\%
over standard attention.

\begin{table}[t]
\caption{Accuracy on FSNS test set.}
\label{tab:results}
\begin{center}
\begin{tabular}{| c | c | c |}  \hline
  CNN & Attention & Accuracy \\ \hline
Smith et al. \cite{smith2016end} & NA & 72.46\% \\ 
Inception-v2 & Standard & 80.7\% \\  
Inception-v2 & Location & 81.8\% \\
Inception-v3 & Standard & 83.1\% \\  
Inception-v3 & Location & 84.0\% \\
Inception-resnet-v2 & Standard & 83.3\% \\  
Inception-resnet-v2 & Location & {\bf 84.2}\% \\
\hline
\end{tabular}
\end{center}
\end{table}

\subsection{The effect of depth on FSNS}
\label{sec:cuts}

The main computational bottleneck in our model is the CNN-based
feature extractor. To see how accuracy and speed vary as a function of
the depth of the CNN, we considered ``cutting'' standard models at
different layers, and then using these as feature extractors.
We evaluate the accuracy of the resulting
trained model (measured as the percentage  of
full sequences predicted correctly),
as well as the speed (measured in milliseconds per single image inference on Tesla K40 GPU).
Table~\ref{tab:inception-v2-results} shows the results for 
inception-v2,
Table~\ref{tab:inception-v3-results}
shows the results for inception-v3,
and Table~\ref{tab:inception-resnet-v2-results}
shows the results for inception-resnet-v2.
Note that these results are using standard attention,
not location-aware attention.

We see that  the accuracy improves for a while,
and then starts to drop as the depth increases. This trend holds for all three models.
We believe the reason for this is that
character recognition does not benefit from the high-level features
that are needed for image classification. Also, the spatial resolution of the image features used as input for attention decreases after every max pooling operation, which limits the precision of the attention mask on a particular character. We don't see any dependency between accuracy and the theoretical receptive field of the neurons in the last convolutional layer, but the effective field of view can be much smaller.

When comparing the different architectures,
we see that inception-resnet-v2 is the most accurate (0.833),
then inception-v3 (0.831),
and finally inception-v2 (0.807).
We chose to use inception-v3 features from the mixed-5d layer for all
the other experiments in this paper,
since this is almost twice as fast as the best inception-resnet-v2 cut,
and has very similar accuracy, it is optimal choice of the architecture for the given computational budget. For the fixed spatial resolution, processing time grows with the depth of the network. After the max pooling layer which decreases the spatial resolution of image features used in attention, we usually observe speed up in the processing time. 

\begin{center}
\begin{table}[t]
  \caption{Performance of different cuts of inception-v2.
    Mixed layers are cuts after the inception block of concatenated
    convolutions.
    }
\label{tab:inception-v2-results}

\begin{tabular}{| c | c | c | c | c | c |}
\hline Inception-v2    & Size            & Acc    & ms/     & Depth & Rec. \\
              layer            & per view     &           & Image &           & Field  \\ \hline \hline
MaxPool\_3a\_3x3    & 19x19x192 & 0.539 & 26       & 3        & 27    \\ \hline
Mixed\_3b                 & 19x19x256 & 0.777 & 31       & 6        & 59    \\ \hline
Mixed\_3c                 & 19x19x320 & 0.803 & 37       & 9        & 91    \\ \hline
Mixed\_4a                 & 10x10x576 & 0.765 & 34       & 12      & 155   \\ \hline
Mixed\_4b                 & 10x10x576 & 0.789 & 37       & 15      & 219   \\ \hline
Mixed\_4c                 & 10x10x576 & 0.805 & 38       & 18      & 283   \\ \hline
Mixed\_4d                 & 10x10x576 & 0.804 & 41       & 21      & 347   \\ \hline
Mixed\_4e                 & 10x10x576 & \textbf{0.807} & 44 & 24 & 411   \\ \hline
Mixed\_5a                 & 5x5x1024   & 0.791 & 43       & 27      & 539   \\ \hline
Mixed\_5b                 & 5x5x1024   & 0.760 & 45       & 30      & 667   \\ \hline
Mixed\_5c                 & 5x5x1024   & 0.792 & 47       & 33      & 795   \\ \hline
\end{tabular}
\end{table}
\end{center}

\begin{table}[t]
  \caption{Performance of different cuts of inception-v3.}
\label{tab:inception-v3-results}
\begin{center}
\begin{tabular}{| c | c | c | c | c | c |}
\hline Inception-v3  & Size           & Acc    & ms/     & Depth & Rec. \\
       layer                & per view     &           & Image &           & Field  \\ \hline \hline
MaxPool\_3a\_3x3 & 35x35x64   & 0.157 & 26       & 3        & 15 \\ \hline
Conv2d\_3b\_1x1   & 35x35x80   & 0.541 & 31      & 4         & 15 \\ \hline
Conv2d\_4a\_3x3   & 33x33x192 & 0.674 & 37      & 5         & 23 \\ \hline
MaxPool\_5a\_3x3 & 16x16x192 & 0.712 & 35      & 5         & 31 \\ \hline
Mixed\_5b              & 16x16x256 & 0.818 & 29      & 8         & 63 \\ \hline
Mixed\_5c              & 16x16x288 & 0.816 & 33      & 11        & 95 \\ \hline
Mixed\_5d              & 16x16x288 & \textbf{0.831} & 36 & 14 & 127 \\ \hline
Mixed\_6a              & 7x7x768     & 0.822 & 34      & 17       & 159 \\ \hline
Mixed\_6b              & 7x7x768     & 0.804 & 37      & 22       & 351 \\ \hline
Mixed\_6c              & 7x7x768     & 0.822 & 40      & 27       & 543\\ \hline
Mixed\_6d              & 7x7x768     & 0.820 & 42      & 32       & 735\\ \hline
Mixed\_6e              & 7x7x768     & 0.826 & 45      & 37       & 927 \\ \hline
Mixed\_7a              & 3x3x1280   & 0.674 & 45      & 41       & 1033 \\ \hline
Mixed\_7b              & 3x3x2048   & 0.802 & 48      & 44       & 1161 \\ \hline
Mixed\_7c              & 3x3x2048   & 0.810 & 51      & 47       & 1289 \\ \hline
\end{tabular}
\end{center}
\end{table}

\begin{table}[t]
\caption{Performance of different cuts of inception-resnet-v2. We only report results from layer Mixed\_5b and above as cuts for lower layers are identical to the inception-v3 network.}
\label{tab:inception-resnet-v2-results}
\begin{center}
\begin{tabular}{| c | c | c | c | c | c |}
\hline Inception-resnet-v2 & Size            & Acc    & ms/     & Depth & Rec. \\
       layer                           & per view     &           & Image &           & Field \\ \hline \hline
Mixed\_5b                         & 16x16x320 & 0.818 & 43       & 8        & 63 \\ \hline
Mixed\_5b + 5                   & 16x16x320 & 0.822 & 57       & 28      & 223 \\ \hline
Mixed\_5b + 10                 & 16x16x320 & 0.824 & 66       & 48      & 383 \\ \hline
Mixed\_6a                         & 7x7x1088   & \textbf{0.833}   & 70 & 51 & 415\\ \hline
Mixed\_6a + 5                   & 7x7x1088   & 0.827 & 90       & 71      & 895 \\ \hline
Mixed\_6a + 10                 & 7x7x1088   & 0.817 & 125     & 91      & 1375 \\ \hline
Mixed\_6a + 15                 & 7x7x1088   & 0.829 & 136     & 111     & 1855 \\ \hline
Mixed\_6a + 20                 & 7x7x1088   & 0.824 & 158     & 131    & 2335 \\ \hline
Mixed\_7a                         & 3x3x2080   & 0.819 & 307     & 134    & 2399 \\ \hline
Mixed\_7a + 5                   & 3x3x2080   & 0.823 & 406     & 154    & 2719 \\ \hline
Mixed\_7a + 10                 & 3x3x2080   & 0.825 & 512     & 174    & 3039 \\ \hline
Conv2d\_7b\_1x1             & 3x3x1536   & 0.832 & 546      &   175  & 3039 \\ \hline
\end{tabular}
\end{center}
\end{table}

\subsection{Visualization of attention on FSNS}

To understand the behavior of our model,  we use a visualization
procedure similar to one proposed by \cite{simonyan2013deep}. For
every predicted character $k$, we compute the partial derivative of
its logit $a_{k}$ with respect to the input image $x$. We can then
compute the saliency of pixel $(i,j)$ using the formula $v_{i,j} =
\left\lVert \frac{\partial a_{k}}{\partial x_{i,j,:}} \right\lVert_2$, 
where we compute the L2 norm of the derivative across all the color
channels. To get a less noisy saliency map, we create 16 slightly
perturbed versions of the image (by adding a small amount of Gaussian
noise) and then average the results. Additionally, we visualize the
attention map $\alpha_t$ by upsampling with nearest-neighbor
interpolation to the size of the input. 

\begin{figure}
\begin{center}
\includegraphics[scale=0.25]{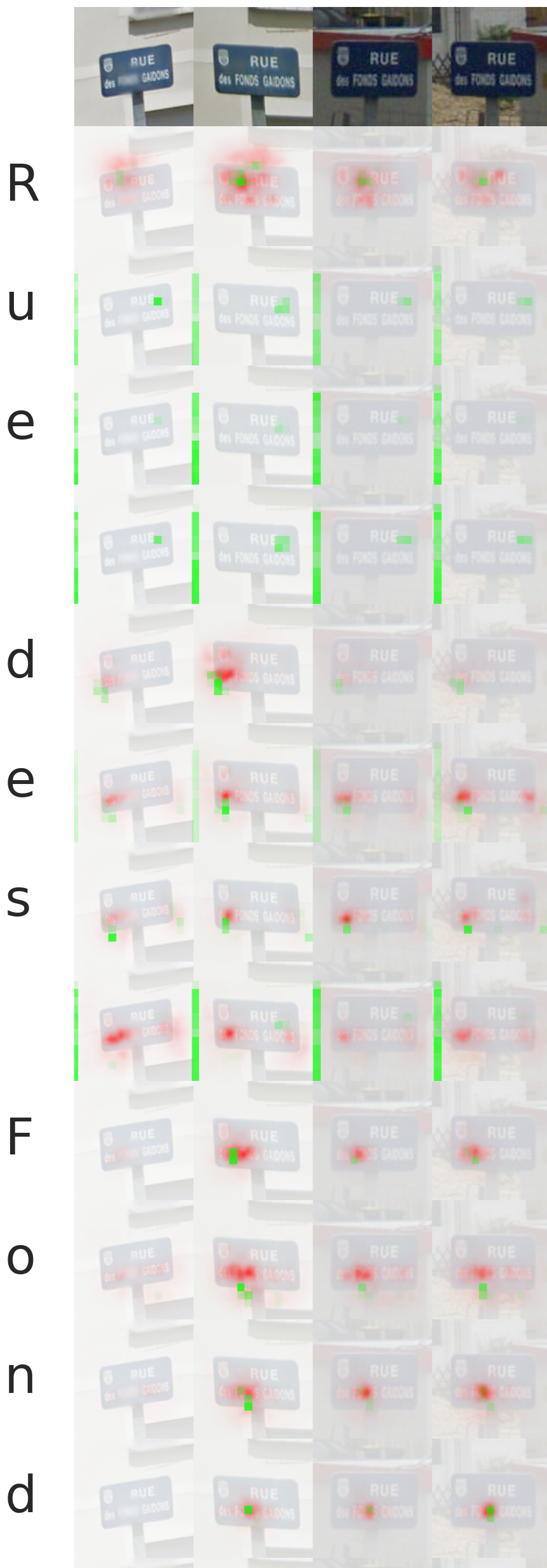}  
\caption{Visualization of saliency maps (in red) and attention masks
  (in green) on an FSNS image.
  The model correctly predicts the string ``Rue des Fonds Gadons''; we
  just show the first 12 steps.
}
\label{fig:saliency}
\end{center}
\end{figure}

A visualization example is shown in Figure~\ref{fig:saliency}. The red
color represents the saliency map, and the green color represents the
attention map. On the first time step, the attention map highlights the
letter 'R,' which is the start of the name. Since 'Rue' is the most
common beginning of a street name in this dataset, the model can
predict the string 'Rue' without paying attention to the image for the
next three iterations. This phenomenon is represented by the vertical
green bars, which show that the model is just paying attention to the
edge of the image, which has no informative content. We also see that
the saliency map is spatially diffused in the first step since any of the
letters 'R', 'u' or 'e' give evidence in support of 'Rue'; for the
remaining three iterations, the saliency map is zero everywhere. 

On the 9th step, the model correctly emits the letter 'F,' which is
the start of the word 'Fonds'. Note how this word is blurred out in
the first view (top left image): this kind of blurring occurs on some words in FSNS,
due to mistakes in the license plate or face detection systems.
Consequently, the attention and saliency maps are zero for the first
view (left column) when processing 'Fonds'.

\subsection{Error Analysis on FSNS}

We analyze 100 randomly sampled wrong predictions to understand the
weaknesses of our model better. 48\% of the ``errors'' is
due to the incorrect ground truth. Table~\ref{tab:errors} gives a more
detailed breakdown. The most common error is due to  
the wrong accent over the letter e (it should be either \'e or \`e);
interestingly, this is also the most common mistake in the ground truth transcription.

\begin{table}[t]
\caption{Breakdown of error types on FSNS.}
\label{tab:errors}
\begin{center}
\begin{tabular}{| c | c |}
\hline Error type                           & Percent \\ \hline \hline
Wrong ground truth                          & 48 \\ \hline
Wrong / Added / Missing accent over \textit{e}       & 17 \\ \hline
Wrong single letter inside the word         & 9 \\ \hline
Wrong single letter at the                  & 8 \\
beginning / end of word                     &   \\ \hline
Added / Missing hyphen (-)                  & 7 \\ \hline
Wrong full word                             & 6 \\ \hline
Read from the wrong view                    & 3 \\ \hline
Wrong / Added / Missing accent              & 2 \\
over different letter than e                &   \\ \hline
\end{tabular}
\end{center}
\end{table}

\begin{figure*}
  \begin{center}
    \subfigure[Confused by font. Pred = 'Avenue Georges Frere', GT =
      'Avenue General Frere'.]
    {\includegraphics[scale=0.5]{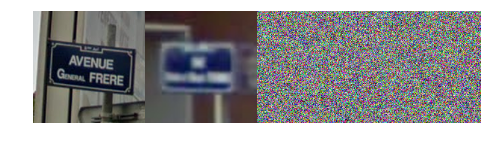}}
    \subfigure[hang][Read text from the wrong
        view. Pred='Boulevard des Talus', GT='Boulevard Charles'.]
              {\includegraphics[scale=0.5]{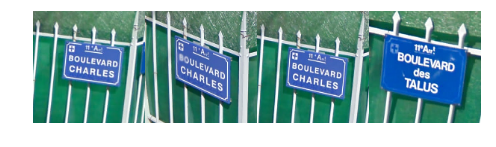}}
    \subfigure[hang][Confusion due to scratched  letter, which looks
      like 'J', but model uses its prior to produce 'O'. Pred='Impasse des
      Jorf\`evres', GT='Impasse des Orf\'evres'.]
              {\includegraphics[scale=0.5]{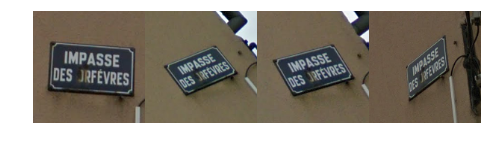}}
    \subfigure[hang][The model has better language prior than the human annotator. Pred='Avenue des Erables', Wrong GT='Avenue des Enadles'.]
              {\includegraphics[scale=0.5]{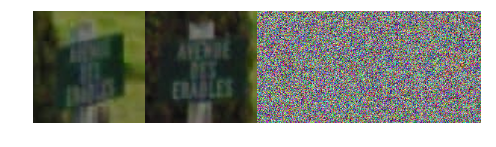}}
  \end{center}
  \caption{Examples of FSNS signs
    where our prediction (Pred.) differs from ground truth (GT).}
  \label{fig:FSNSresults}

  \begin{center}
    \subfigure{
    \includegraphics[width=0.41\hsize]{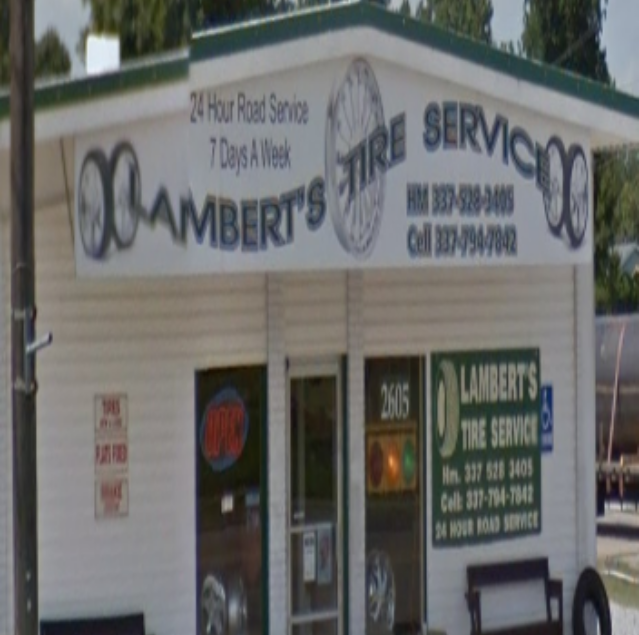}
    \label{fig:bizname_image}
  }
  \subfigure{
    \includegraphics[width=0.41\hsize]{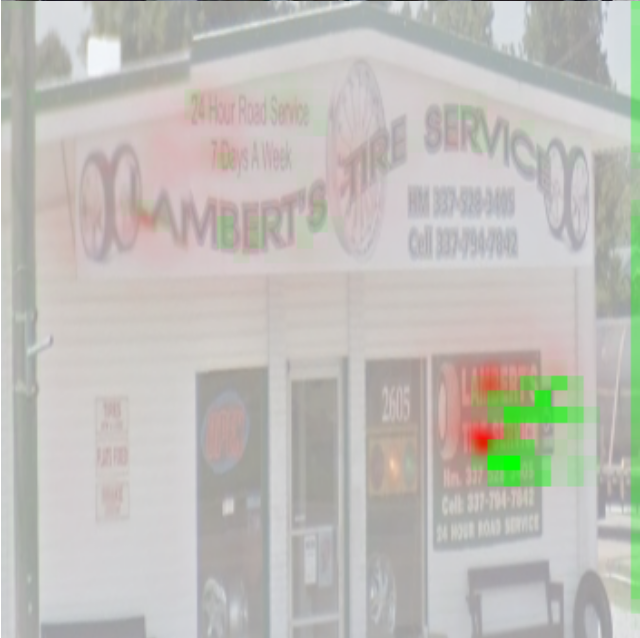}
    \label{fig:bizname_res}
  }
  \end{center}
  \caption{Left: An example image from the internal Business Names dataset.
    The ground truth string is ``Lambert's Tire Service'',
    which is correctly predicted by the model.
    This name is written on the top of the store in a ``wavy'' font, but also on the right-hand side (green sign) in a more standard horizontal font.
    Right: Visualization of the time-averaged saliency maps (in red) and
    attention masks (in green). Notice how they focus on the location
    that contains the store name.
}
  \label{fig:bizname}
\end{figure*}

In Figure~\ref{fig:FSNSresults}, we show some of the test cases where
our model has a different prediction than the ground truth. 
(In the last example, this is due to an error in the ground truth.)

\subsection{Results on internal datasets}
We now report results on a new internal dataset.
Since this dataset is not public,
we just show qualitative results;
the purpose is to
show how the very same model can be used for extracting information
from many different kinds of street signs.

Figure~\ref{fig:bizname} shows an example of our system applied to
an image from the Business Names dataset. Notice how the name of the
business, ``Lambert's Tire Service'', appears in two locations: at the
top, and at the bottom right. The model chooses to attend to the
latter, perhaps because the font is more standard, and the writing is
horizontal and not ``wavy''. 

\begin{figure}
\begin{center}
\includegraphics[height=3.6in]{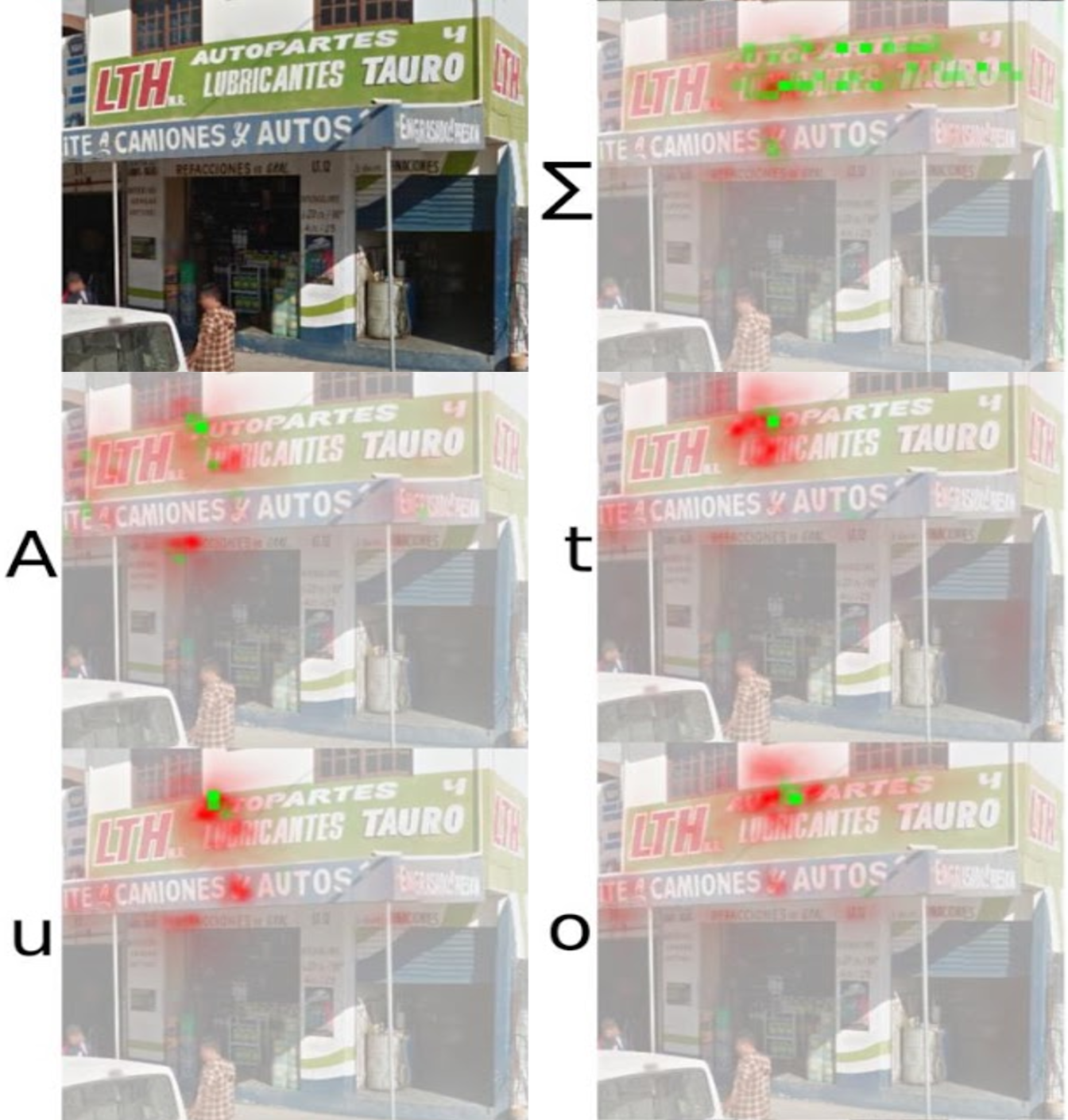}
\caption{Visualization of saliency maps (in red) and attention masks
  (in green) on a Business Names image.
  The model correctly predicts the string ``Autopartes Lubricantes Tauro'';
  we just show the first few steps.
}
\label{fig:bizname_heatmap}
\end{center}
\end{figure}

Figure~\ref{fig:bizname_heatmap} shows another example.
This time, we see that the attention maps scans left to right,
to read ``Autopartes'',
but has to jump down a line and all the way to the left
(like the carriage return action of a mechanical typewriter)
to read the second line of text, ``Lubricantes Tauro''.

 \section{Conclusions and future work}

We have presented an end-to-end approach for scene text recognition
which gives state-of-the-art results on the challenging FSNS dataset,
and an internal dataset. Our novel attention mechanism allows us to
extract structured text information by reading only the interesting
parts of the whole image. 
\\ \\ \\ \\
In the future, we would like to investigate more sophisticated ways of
training the RNN, such as scheduled sampling \cite{Bengio2015} or
hybrid ML/RL methods \cite{Norouzi2016}. We would also like to extend
the system to full structured extraction of business information from
storefronts. \\ \\
\section*{Acknowledgment}
\emph{
The authors would like to thank Sergio Guadarrama, Alex Lacoste, Ray Smith, Quoc Le, Christian Szegedy, Oriol Vinyals, Ilya Sutskever, Sergey Ioffe, Vincent Vanhoucke and Sacha Arnoud for valuable discussions and TensorFlow support. 
}

\bibliographystyle{IEEEtran}
\bibliography{bib}

%


\newpage




%

\end{document}